\documentclass[conference]{IEEEtran}
\usepackage{cite}
\usepackage{amsmath,amssymb,amsfonts}
\usepackage{algorithmic}
\usepackage{algorithm}
\usepackage{graphicx}
\usepackage{textcomp}
\usepackage[table]{xcolor}
\usepackage{booktabs}
\usepackage{multirow}
\usepackage{balance}
\usepackage[utf8]{inputenc}
\usepackage{tikz}
\usepackage{pgfplots}
\usepackage{pgf-pie}

\newcommand{\authcell}[4]{\begin{tabular}{@{}c@{}}\textbf{#1}\\ \textit{#2}\\ \textit{#3}\\ #4\end{tabular}}

\begin{document}

\title{Hierarchical Dual-Head Model for Suicide Risk Assessment via MentalRoBERTa}
\author{%
{\small
\begin{tabular*}{\textwidth}{@{\extracolsep{\fill}} c c c}
\authcell{Chang YANG\textsuperscript{*}}{Department of Informatics}{King's College London}{London, United Kingdom\\ chang.2.yang@kcl.ac.uk} &
\authcell{Ziyi WANG\textsuperscript{*}}{School of Artificial Intelligence}{Beijing University of Posts and Telecommunications}{Beijing, China\\ ziyiwang2003@bupt.edu.cn} &
\authcell{Wangfeng TAN\textsuperscript{*}}{School of Computer Science and Technology}{Anhui University}{An Hui, China\\ wftan2004@gmail.com} \\[4em]
\authcell{Zhiting TAN}{School of Software}{Nankai University}{Tianjin, China\\ 2120250823@mail.nankai.edu.cn} &
\authcell{Changrui JI}{Department of Digital Humanities}{King's College London}{London, United Kingdom\\ changrui.ji@kcl.ac.uk} &
\authcell{Zhiming ZHOU}{School of Arificial Intelligence}{Anhui University}{Anhui, China\\ wa23301159@stu.ahu.edu.cn}
\end{tabular*}
}
}

\maketitle

\begin{abstract}    Social media platforms have become important sources for identifying suicide risk, but automated detection systems face multiple challenges including severe class imbalance, temporal complexity in posting patterns, and the dual nature of risk levels as both ordinal and categorical. This paper proposes a hierarchical dual-head neural network based on MentalRoBERTa for suicide risk classification into four levels: indicator, ideation, behavior, and attempt. The model employs two complementary prediction heads operating on a shared sequence representation: a CORAL (Consistent Rank Logits) head that preserves ordinal relationships between risk levels, and a standard classification head that enables flexible categorical distinctions. A 3-layer Transformer encoder with 8-head multi-head attention models temporal dependencies across post sequences, while explicit time interval embeddings capture posting behavior dynamics. The model is trained with a combined loss function (0.5 CORAL + 0.3 Cross-Entropy + 0.2 Focal Loss) that simultaneously addresses ordinal structure preservation, overconfidence reduction, and class imbalance. To improve computational efficiency, we freeze the first 6 layers (50\%) of MentalRoBERTa and employ mixed-precision training. The model is evaluated using 5-fold stratified cross-validation with macro F1 score as the primary metric.
    \end{abstract}
    

    \section{Introduction}
    
    Suicide is a major global public health issue, with over 700,000 deaths reported annually by the World Health Organization \cite{world2025suicide}. Early identification of at-risk individuals is critical for timely intervention and prevention. In recent years, social media platforms have become venues where individuals express mental distress and suicidal thoughts, often before seeking professional help \cite{coppersmith2018natural}. Reddit, in particular, hosts numerous mental health communities (e.g., r/depression, r/SuicideWatch) where users discuss anxiety and suicidal ideation. This linguistic data provides opportunities for developing automated risk detection systems based on natural language processing.
    
    However, suicide risk detection from social media text presents several technical challenges. First, the data exhibits severe class imbalance. High-risk samples (behavior and attempt categories) are significantly less frequent than low-risk samples (indicator and ideation), causing standard classifiers to be biased toward majority classes \cite{ji2018supervised}. Second, risk levels have dual characteristics: they follow an ordinal progression (indicator $<$ ideation $<$ behavior $<$ attempt) in terms of severity, but also have categorical distinctions that require different clinical intervention strategies. Existing models typically employ either ordinal regression methods \textit{or} standard classification, but not both, thus missing complementary information \cite{burnap2015machine}. Third, suicide risk evolves over time, reflected in temporal patterns across multiple posts. Capturing these temporal dependencies requires sequence modeling beyond simple bag-of-words or single-post analysis \cite{tadesse2019detection}. A sudden increase in posting frequency or shift in sentiment may indicate crisis escalation. Fourth, while large pre-trained language models like BERT and RoBERTa offer strong semantic understanding \cite{devlin2019bert}, they are computationally expensive to fine-tune on domain-specific mental health data, requiring efficient training strategies.

    Recent work has explored deep learning approaches for suicide risk detection, including LSTM-based sequence models \cite{roy2020machine} and BERT-based classifiers. However, most existing methods focus on single-head architectures with simple loss functions like cross-entropy. Few approaches explicitly model both the ordinal structure and categorical distinctions of risk levels, or combine multiple complementary objectives to address class imbalance and temporal evolution simultaneously.

    This paper proposes a hierarchical dual-head model that addresses these challenges through several key innovations. First, this paper employs a dual-head architecture combining CORAL ordinal regression \cite{cao2020rank} and standard classification. Both heads operate on the same learned sequence representation, allowing their outputs to be ensembled at inference time for more robust predictions. The CORAL head is trained with a single CORAL loss to preserve ordinal relationships, while the classification head is optimized with a combined loss function incorporating both cross-entropy with label smoothing and focal loss to address overconfidence and class imbalance simultaneously. Second, this paper uses a 3-layer Transformer encoder to model inter-post dependencies and temporal evolution. Time intervals between consecutive posts are explicitly encoded via a learned embedding network with LayerNorm, capturing posting frequency patterns that signal mental state changes. Third, this paper optimizes a weighted combination of three losses: CORAL loss for preserving ordinal structure, cross-entropy with label smoothing to prevent overconfident predictions, and focal loss to emphasize minority classes and hard samples. The fixed weights (0.5, 0.3, 0.2) balance these complementary objectives. Fourth, this paper employs MentalRoBERTa, a RoBERTa model pre-trained on Reddit mental health texts, and freezes 50\% of its parameters to reduce computational cost while maintaining domain knowledge. This paper also applies mixed-precision training to reduce memory usage. Finally, the model supports optional statistical feature extraction, including post length patterns and time interval statistics, which can be fused with the text representation when enabled.
    
    The main contributions of this work are:
    \begin{itemize}
    \item A dual-head architecture combining CORAL ordinal regression and classification for suicide risk prediction, leveraging complementary prediction paradigms through ensemble.
    \item Temporal modeling through a 3-layer Transformer encoder with explicit time interval embeddings to capture posting behavior dynamics and mental state evolution.
    \item A combined loss function (CORAL + Cross-Entropy + Focal) addressing ordinal structure preservation, overconfidence reduction, and class imbalance simultaneously with fixed weights.
    \item Efficient training strategy with partial RoBERTa freezing (6 layers, 50\% parameters), differentiated learning rates, and mixed-precision computation.
    \item Model evaluation framework using 5-fold stratified cross-validation with macro F1 score as the primary performance metric.
    \end{itemize}
    
    \section{Methodology}
    
    \subsection{Problem Formulation}

    We formulate suicide risk assessment as predicting a user's risk level $y \in \{0, 1, 2, 3\}$ from their chronologically ordered sequence of $N$ posts $\mathcal{P}_u = \{p_1, p_2, \ldots, p_N\}$ with corresponding timestamps $\mathcal{T}_u = \{t_1, t_2, \ldots, t_N\}$. Each post $p_i$ is a text string, and each timestamp $t_i$ is a Unix epoch time. The four risk levels are defined as: (0) indicator - expression of suicide-related indicators or warning signs; (1) ideation - expression of suicidal thoughts without specific plans; (2) behavior - planning or preparatory actions for suicide; (3) attempt - explicit intent to attempt or history of past attempts. These labels exhibit ordinal relationships ($0 < 1 < 2 < 3$) reflecting increasing severity, while also requiring categorical distinction for clinical intervention design.
    
    In practice, we construct sequences by grouping consecutive posts from the same user. Each sequence consists of 5 consecutive posts with their corresponding timestamps, enabling temporal analysis of posting patterns. The risk level label is determined by the 6th post immediately following the sequence, which serves as the ground truth for predicting the user's current risk state.

    \subsection{Model Architecture}
    
    \begin{figure*}[t]
    \centering
\includegraphics[width=0.85\textwidth]{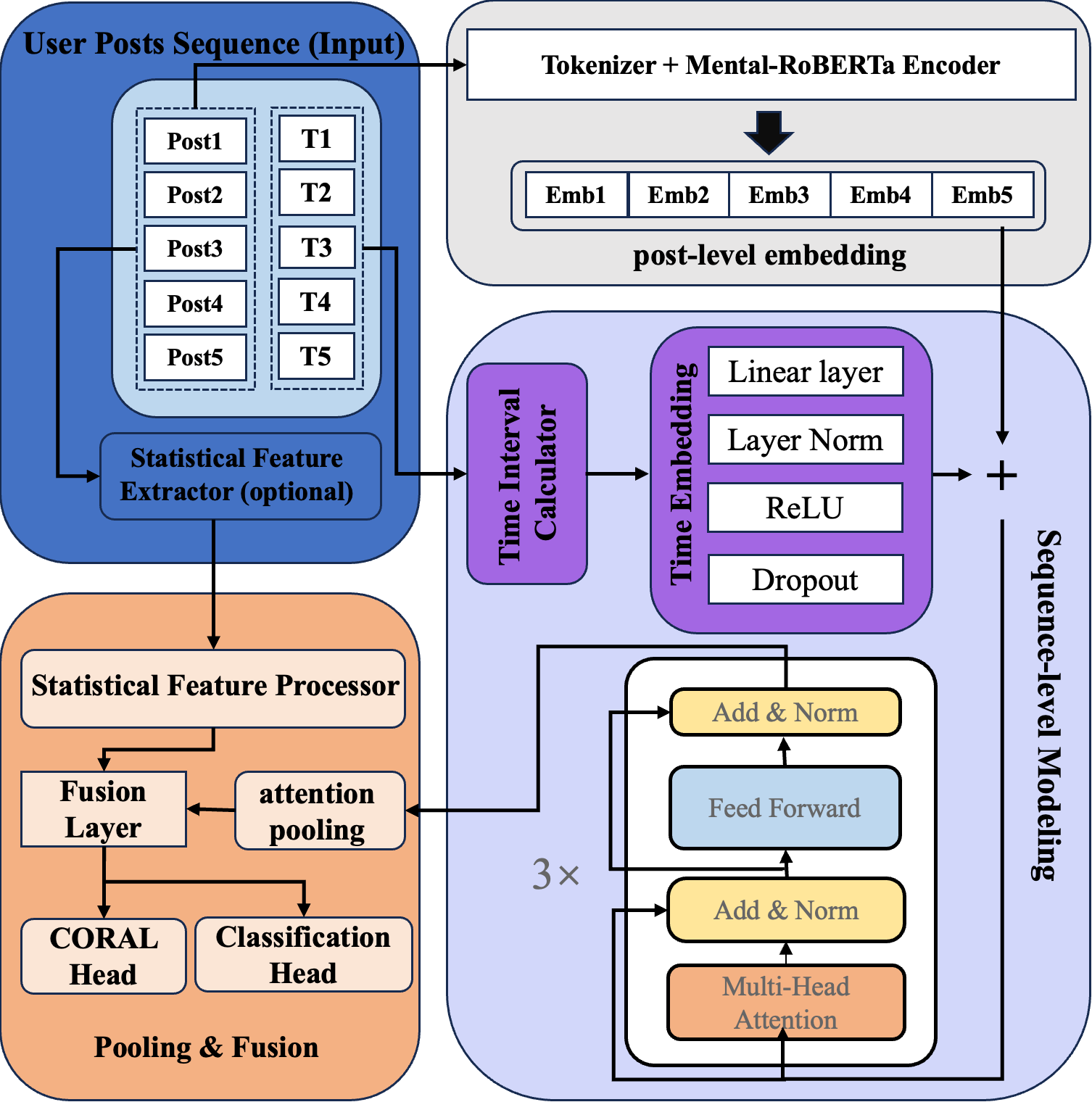}
    \caption{Overview of the hierarchical dual-head architecture.}
    \label{fig:architecture}
    \end{figure*}
    
    Figure \ref{fig:architecture} shows the model architecture. The model operates at two hierarchical levels: post-level encoding processes individual posts to obtain semantic representations, and sequence-level aggregation models the temporal sequence of posts to produce a fixed sequence representation. The final prediction is made by two complementary heads operating on this sequence representation. The architecture consists of the following components:
    
    1) \textit{Post-Level Encoding with MentalRoBERTa:} This paper uses MentalRoBERTa \cite{ji2021mentalbert} as the base encoder. MentalRoBERTa is a RoBERTa-base model (12 layers, 768 hidden dimensions, 125M parameters) that has been pre-trained on a large corpus of Reddit posts from mental health-related subreddits. This domain adaptation provides better semantic understanding of mental health language compared to generic RoBERTa. Each post $p_i$ is tokenized using the RoBERTa tokenizer with maximum sequence length 512. The tokenizer converts text into subword tokens and adds special tokens [CLS] and [SEP]. Posts shorter than 512 tokens are padded with [PAD] tokens. The tokenized sequence is fed into MentalRoBERTa, which outputs contextualized representations for all tokens. This paper extracts the [CLS] token embedding as the post representation: $\mathbf{e}_i \in \mathbb{R}^{768}$. To reduce computational cost and prevent catastrophic forgetting of pre-trained knowledge, this paper employs partial parameter freezing. Specifically, this paper freezes the embedding layer and the first 6 encoder layers (out of 12 total), which accounts for 50\% of RoBERTa's parameters. Only the upper 6 encoder layers are fine-tuned.
    
    2) \textit{Temporal Embedding:} Time intervals between consecutive posts provide important signals about mental state dynamics. For each pair of consecutive posts, this paper computes the time interval $\Delta t_i = t_{i+1} - t_i$ in days. To prevent extreme outliers from dominating the embedding space, this paper caps the interval at 365 days: $\Delta t_i = \min(t_{i+1} - t_i, 365)$. For the first post, this paper sets $\Delta t_0 = 0$. Each interval is then processed through a temporal embedding network:
    \begin{equation}
    \mathbf{e}^{\text{time}}_i = \text{Dropout}(\text{ReLU}(\text{LayerNorm}(\mathbf{W}_t \Delta t_i + \mathbf{b}_t)))
    \end{equation}
    where $\mathbf{W}_t \in \mathbb{R}^{768 \times 1}$ and $\mathbf{b}_t \in \mathbb{R}^{768}$ are learnable parameters. The temporal embedding is added to the post embedding: $\tilde{\mathbf{e}}_i = \mathbf{e}_i + \mathbf{e}^{\text{time}}_i$.
    
    3) \textit{Sequence-Level Transformer Encoder:} The temporally-enhanced embeddings are processed by a 3-layer Transformer encoder \cite{vaswani2017attention}. Each layer consists of: (1) 8-head multi-head self-attention with head dimension 96, (2) residual connection with layer normalization, (3) position-wise feed-forward network with hidden dimension 3072 and GELU activation. The Transformer encoder produces contextualized representations $\mathbf{S} = [\mathbf{s}_1, \ldots, \mathbf{s}_5] \in \mathbb{R}^{5 \times 768}$. Each $\mathbf{s}_i$ encodes not only the content of post $i$, but also its relationship to all other posts.
    
    4) \textit{Attention Pooling:} To aggregate the sequence into a fixed sequence representation, this paper employs 4-head multi-head attention with a learnable query vector $\mathbf{q} \in \mathbb{R}^{768}$:
    \begin{equation}
    \mathbf{u} = \text{MultiHeadAttn}(\mathbf{q}, \mathbf{S}, \mathbf{S})
    \end{equation}
    The attention weights are learned to focus on the most risk-relevant posts. For padding posts, this paper applies a key padding mask. The output is a fixed sequence representation $\mathbf{u} \in \mathbb{R}^{768}$.
    
    5) \textit{Optional Statistical Features:} The model optionally extracts behavioral patterns from the post sequence, including post length statistics (mean, std, min, max word counts) and time interval statistics. These features are processed through a two-layer MLP producing $\mathbf{f}_{\text{stat}} \in \mathbb{R}^{64}$. When enabled, they are concatenated with $\mathbf{u}$ and projected: $\mathbf{u}_{\text{fused}} = \text{Linear}([\mathbf{u}; \mathbf{f}_{\text{stat}}])$.
    
    6) \textit{Dual Prediction Heads:} Two complementary heads operate on the sequence representation. The CORAL head learns 3 binary thresholds using shared weight $\mathbf{w}_c \in \mathbb{R}^{768}$ and ordered biases $b_1 < b_2 < b_3$:
    \begin{equation}
    P(y > k \mid \mathbf{u}) = \sigma(\mathbf{w}_c^T \mathbf{u} + b_k), \quad k \in \{0, 1, 2\}
    \end{equation}
    The classification head is a standard linear layer: $\mathbf{z}_{\text{class}} = \mathbf{W}_{\text{class}} \mathbf{u} + \mathbf{b}_{\text{class}} \in \mathbb{R}^4$. The CORAL head is trained with a single CORAL loss to preserve ordinal relationships, while the classification head is optimized with a combined loss function incorporating both cross-entropy with label smoothing and focal loss. At inference, this paper ensembles by averaging: $\mathbf{p}_{\text{final}} = 0.5 \cdot \mathbf{p}_{\text{CORAL}} + 0.5 \cdot \mathbf{p}_{\text{class}}$.
    
    \subsection{Loss Function}
    
    We optimize a weighted combination of three losses:
    \begin{equation}
    \mathcal{L}_{\text{total}} = 0.5 \mathcal{L}_{\text{CORAL}} + 0.3 \mathcal{L}_{\text{CE}} + 0.2 \mathcal{L}_{\text{Focal}}
    \end{equation}
    
    1) \textit{CORAL Loss:} For a sample with true label $y$, this paper constructs binary targets $t_k = \mathbb{1}[y > k]$ for each threshold $k \in \{0, 1, 2\}$:
    \begin{equation}
    \mathcal{L}_{\text{CORAL}} = -\sum_{k=0}^{2} \left[ t_k \log \sigma(z^c_k) + (1-t_k) \log(1 - \sigma(z^c_k)) \right]
    \end{equation}
    where $z^c_k = \mathbf{w}_c^T \mathbf{u} + b_k$.
    
    2) \textit{Cross-Entropy with Label Smoothing:} We apply smoothing factor $\epsilon = 0.1$:
    \begin{equation}
    y_i^{\text{smooth}} = \begin{cases}
    1 - \epsilon & \text{if } i = y \\
    \epsilon / 4 & \text{otherwise}
    \end{cases}
    \end{equation}
    The smoothed cross-entropy loss is:
    \begin{equation}
    \mathcal{L}_{\text{CE}} = -\sum_{i=0}^{3} y_i^{\text{smooth}} \log p_{\text{class},i}
    \end{equation}
    
    3) \textit{Focal Loss:} With focusing parameter $\gamma = 2$ and class weights $\alpha_i$:
    \begin{equation}
    \mathcal{L}_{\text{Focal}} = -\sum_{i=0}^{3} \alpha_i (1-p_{\text{class},i})^{\gamma} y_i \log p_{\text{class},i}
    \end{equation}
    The class weights are computed as inverse class frequencies.
    
    \subsection{Training Strategy}
    \label{subsec:training}
    
    1) \textit{Efficient Fine-Tuning:} We freeze the embedding layer and first 6 encoder layers of MentalRoBERTa, reducing trainable parameters from 125M to approximately 70M. We apply differentiated learning rates: $2 \times 10^{-5}$ for the unfrozen RoBERTa layers and $1 \times 10^{-4}$ for randomly initialized components.
    
    2) \textit{Optimization:} We use the AdamW optimizer with weight decay 0.01\cite{loshchilov2017decoupled}. The learning rate follows a cosine annealing schedule with 10\% warmup steps. We use gradient accumulation over 2 steps with batch size 8 (effective batch size 16). Gradients are clipped to maximum norm 1.0.
    
    3) \textit{Regularization:} This paper applies dropout with rate 0.3 after the temporal embedding, within the Transformer encoder, and after attention pooling. Label smoothing with factor 0.1 provides additional regularization. This paper employs text data augmentation: for each post, this paper randomly applies one of three operations with 50\% probability: (1) random deletion - remove 10\% of words; (2) random swap - swap two random words; (3) synonym replacement - replace 10\% of words with WordNet synonyms.
    
    4) \textit{Mixed-Precision Training:} We employ automatic mixed-precision (AMP) training. Operations that benefit from FP16 precision are computed in FP16, while operations requiring higher precision remain in FP32. This reduces memory consumption by 30-40\% and speeds up training by 20-30\%.
    
    5) \textit{Cross-Validation:} This paper conducts 5-fold stratified cross-validation. The data is split into 5 folds while preserving class distribution. For each fold, this paper trains on 4 folds and validates on the remaining fold. This paper uses early stopping with patience 5 epochs based on validation macro-F1 score.

\section{Data Augmentation}

To address the inherent challenges in suicide risk assessment, such as class imbalance in training datasets, this paper introduces a dedicated data augmentation strategy. This section details the motivation behind augmentation and presents two complementary methods: (1) in-sample enhancement using large language models (LLMs) for generating semantically equivalent variants, and (2) external data acquisition via web crawling combined with LLM-assisted labeling. These approaches aim to enrich the dataset while preserving the ordinal and categorical nuances of risk levels, thereby improving model generalization and robustness.

\subsection{Motivation: Addressing Class Imbalance}

The original dataset exhibits severe class imbalance, with disproportionate representation across the four suicide risk levels: Level 0 (Indicator) comprises 2,480 samples, Level 1 (Ideation) has 3,536 samples, Level 2 (Behavior) includes 1,019 samples, and Level 3 (Attempt) contains only 348 samples. This skew, where high-risk classes (Levels 2 and 3) are underrepresented, can lead to biased models that underperform on minority classes.

To visualize this distribution, this paper presents the following pie chart illustrating the proportion of each class in the original dataset:

\begin{figure}[h]
\centering
\includegraphics[width=0.48\textwidth]{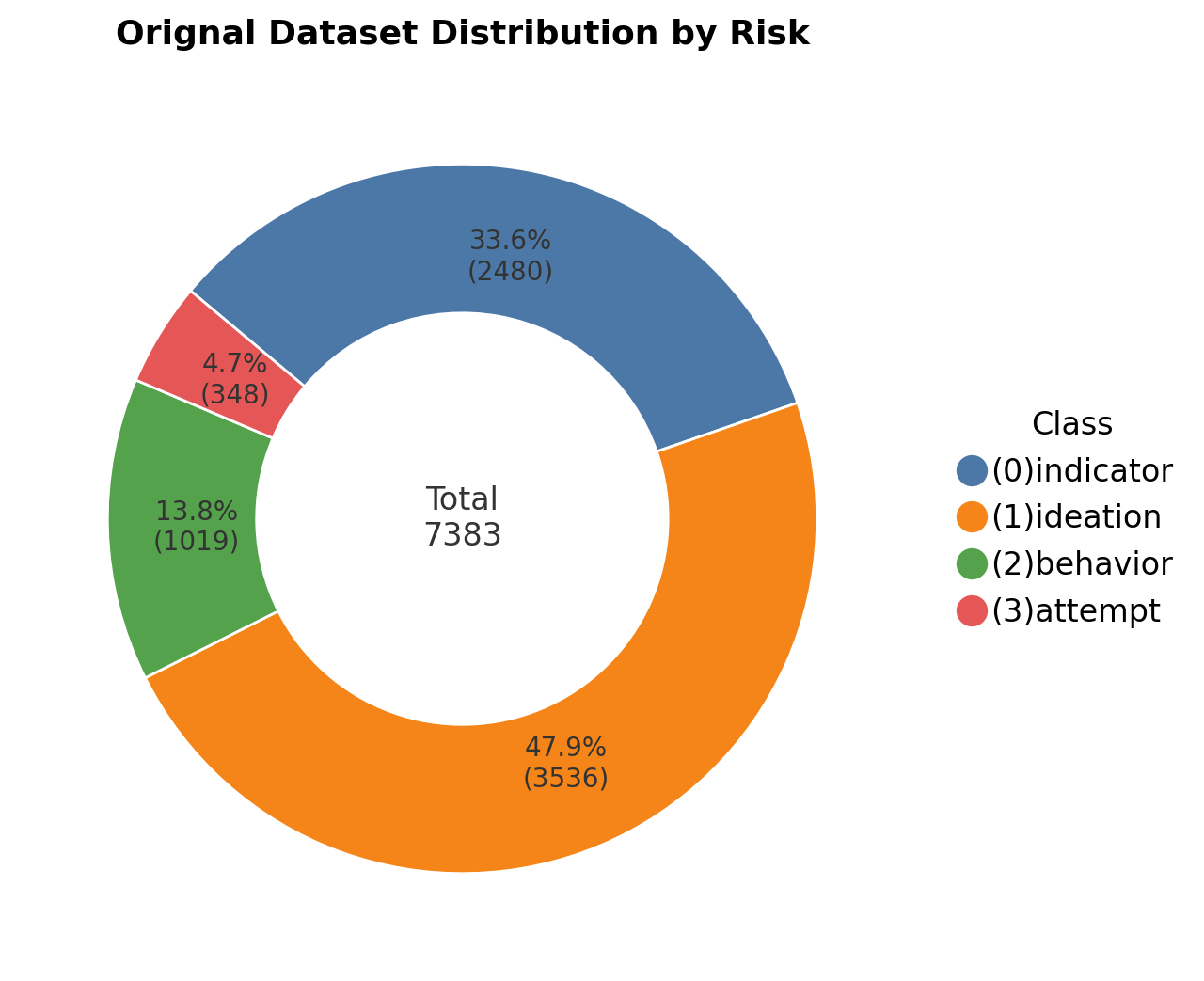}
\caption{Original Dataset Distribution by Risk Level}
\label{fig:orig-dist}
\end{figure}

This visualization underscores the dominance of lower-risk classes (Levels 0 and 1, accounting for over 70\% of samples), motivating targeted augmentation to balance the dataset without introducing noise or semantic drift.

\begin{figure*}[t]
    \centering
    \includegraphics[width=\textwidth]{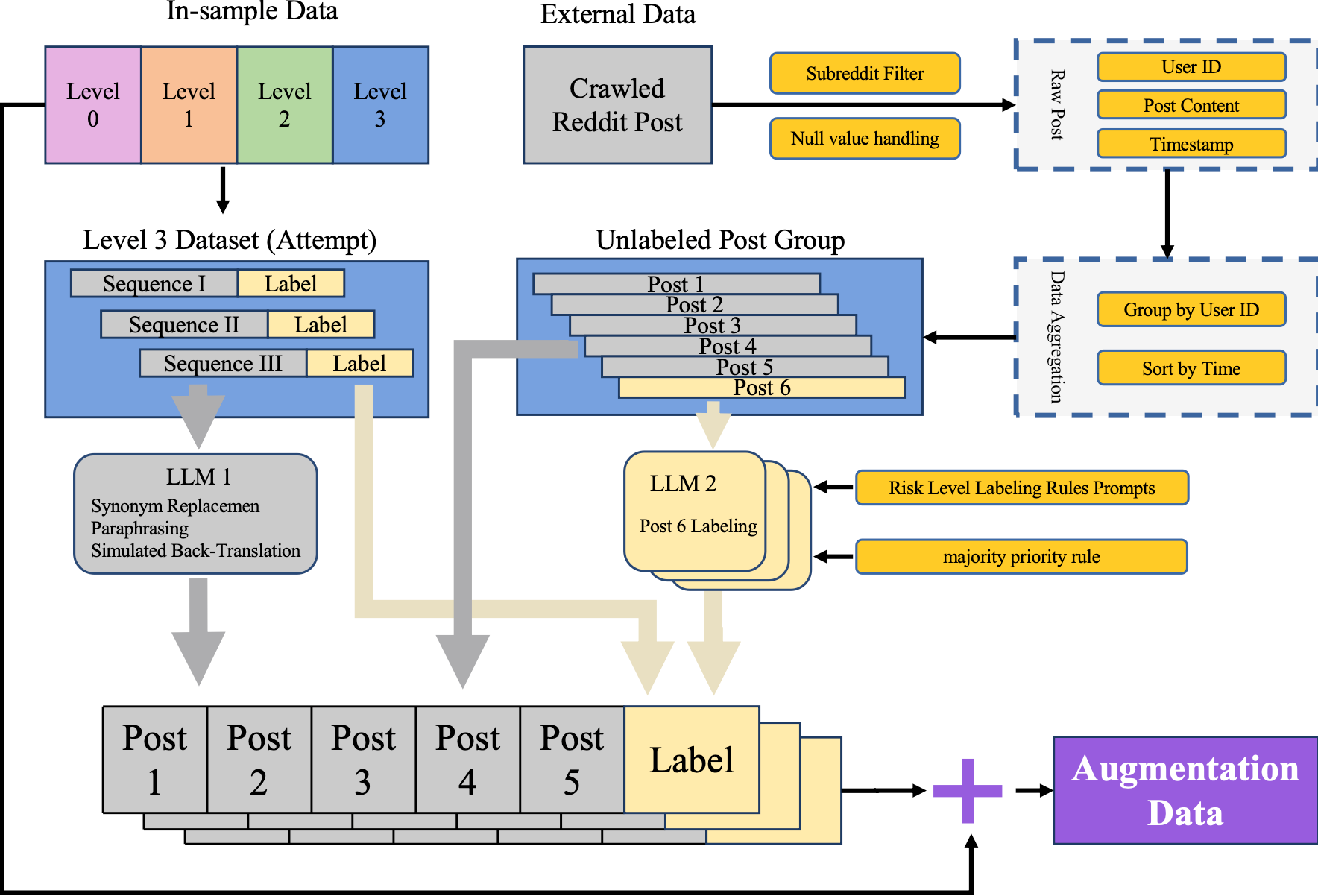}
    \caption{Data Augmentation: Overview of combining in-sample LLM-based augmentation and external data acquisition with LLM-assisted labeling.}
    \label{fig:data_augmentation_pipeline}
    \end{figure*}

\subsection{Method 1: In-Sample Augmentation via LLM-Generated Variants}

To mitigate imbalance internally, this paper leverages a large language model (e.g., GPT-4) prompted to generate augmented variants of existing samples. This method focuses on low-frequency classes (Level 3) by applying two NLP augmentation techniques: paraphrasing with synonym replacement and simulated back-translation. Paraphrasing rephrases sentences using alternative vocabulary and structures while retaining core semantics, whereas simulated back-translation mimics the effect of round-trip translation (e.g., English $\to$ intermediate language $\to$ English) to introduce syntactic diversity without actual bilingual processing.

For each original post in a batch of five (selected from underrepresented classes), the LLM generates exactly three distinct variants per post under strict rules that preserve the original semantics, first-person perspective, and risk level, without adding any new information.

This procedure enforces strict adherence to risk semantics, preventing dilution of high-risk signals. We apply augmentation selectively: for Level 3, generating $3\times$ variants per sample to triple their effective count, while Levels 0 and 1 receive no augmentation to avoid over-representation. Post-augmentation, the dataset is rebalanced via stratified sampling, yielding a more uniform distribution for training.

\subsection{Method 2: External Data Acquisition via Crawling and LLM-Assisted Labeling}

To further diversify the dataset, this paper collects real-world posts from the Reddit subreddit r/suicidewatch using ethical web crawling techniques. This paper retrieves user IDs, post texts, and timestamps, aggregating posts per user and sorting them chronologically. Sequences are constructed by grouping every six consecutive posts into a unlabeled post group, appended with timestamps to capture temporal dynamics. The sixth post in each group serves as a label answer source for labeling the preceding five, using LLM-assisted annotation guided by established psychological standards~\cite{li2022suicide}.

We operationalize four risk levels based on the benchmark framework in~\cite{li2022suicide}, providing clear, rule-based definitions for ideation, behavior (including preparatory acts and plans), attempts, and irrelevant content to guide annotation. To ensure annotation quality and reduce individual model bias, this paper employs multiple large language models (LLMs) in an ensemble approach. Specifically, each sequence is independently annotated by multiple LLMs following the same standardized guidelines. The final label for each sequence is determined through a majority priority rule: the risk level that receives the most votes from the LLM ensemble is selected as the ground-truth label for the sixth post. This voting mechanism enhances annotation reliability and consensus compared to single-model annotation.

These guidelines ensure consistent, rule-based labeling, with Level 4 samples filtered out post-annotation. From approximately 5,000 raw sequences, this paper retains 3,722 high-quality samples after validation, distributed as: Level 0: 718, Level 1: 1,887, Level 2: 931, Level 3: 186.

To quantify the impact, the following bar chart depicts the percentage growth in each class relative to the original dataset:

\begin{figure}[h]
\centering
\includegraphics[width=0.9\linewidth]{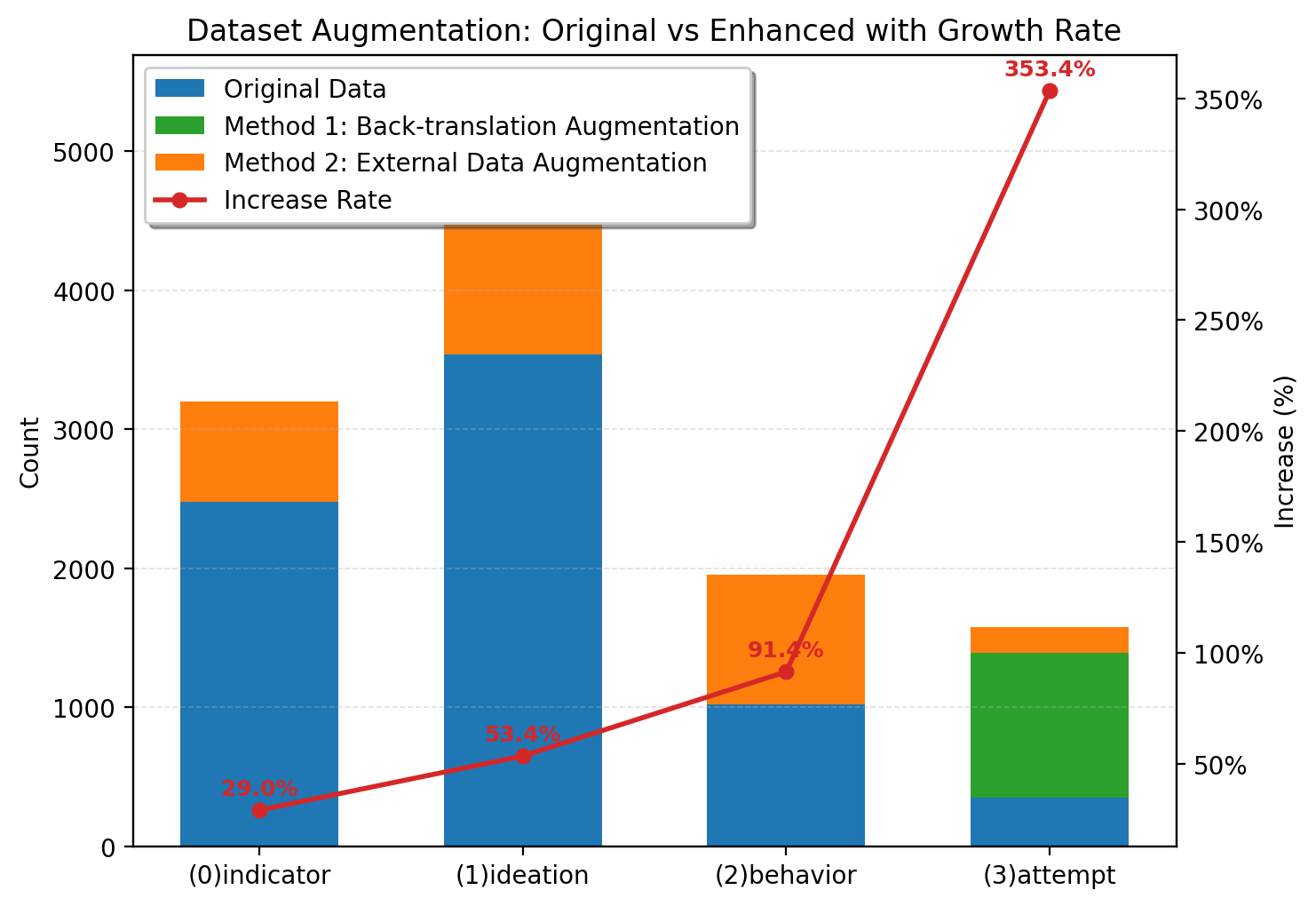}
\caption{Percentage Growth in Dataset Size per Risk Level After Augmentation}
\label{fig:growth-bar}
\end{figure}

Note the substantial relative gains in high-risk classes (e.g., +53\% for Level 3), despite absolute decreases in lower-risk classes due to filtering, resulting in an overall more balanced corpus.

\section{Experiments}
\label{sec:experiments}

\subsection{Experimental Setup}
\label{subsec:exp_setup}

We conduct a series of experiments to evaluate the effectiveness of the proposed hierarchical dual-head model. All experiments are performed using the PyTorch framework. We assess the model on both the original dataset and the augmented dataset described in the previous section. The primary evaluation metrics are Macro F1-score, which is crucial for imbalanced classification tasks; Mean Absolute Error (MAE), to quantify the ordinal prediction error; and Quadratic Weighted Kappa (QWK), to measure the agreement between predicted and true ordinal ranks. The model is trained using the AdamW optimizer with the training strategy detailed in Section~\ref{subsec:training}. All reported results for the main model are derived from a 5-fold stratified cross-validation setup to ensure robustness.

\subsection{Impact of Data Augmentation}
\label{subsec:exp_augmentation}
\begin{figure}[t]
    \centering
    \includegraphics[width=0.9\linewidth]{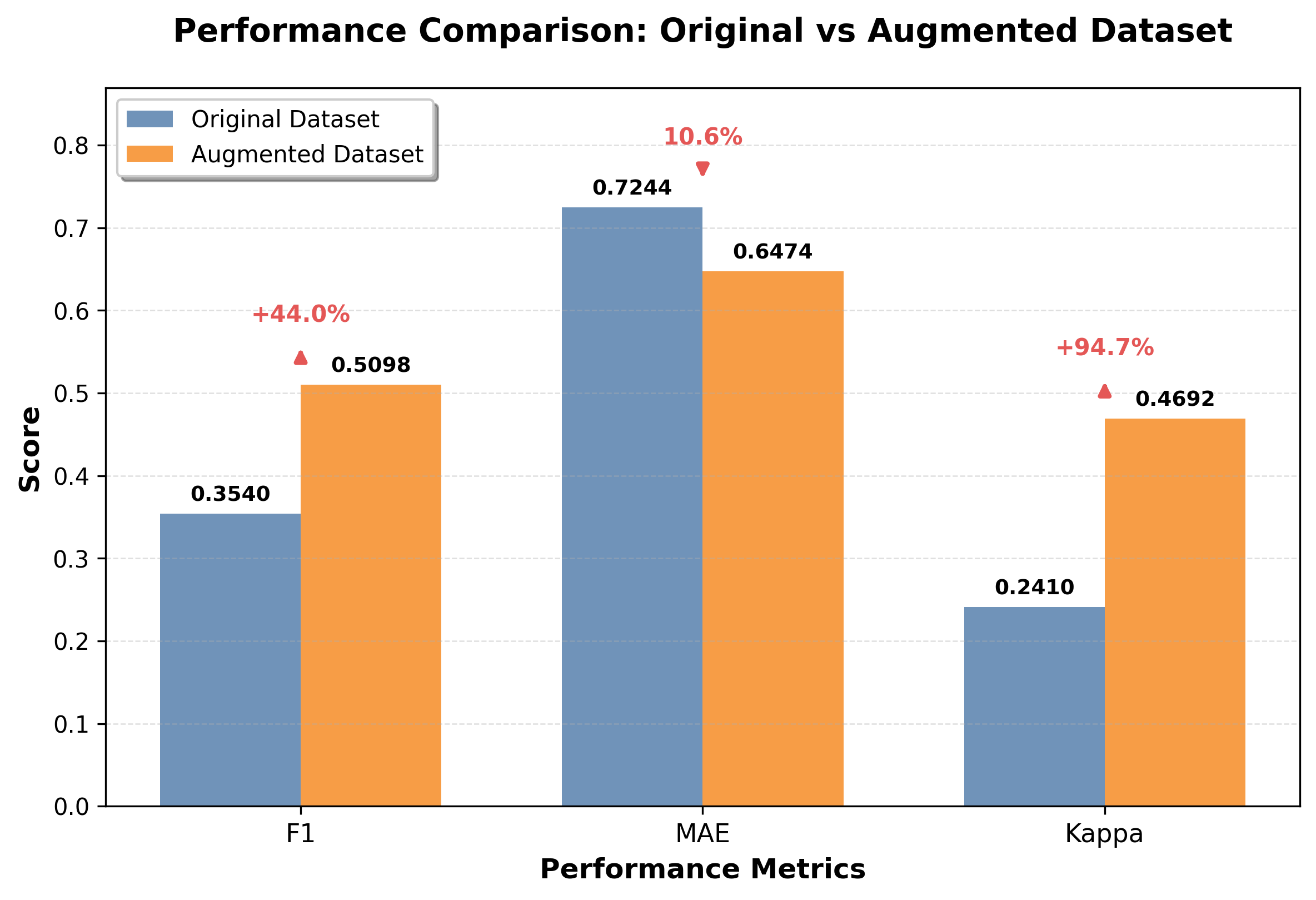} 
    \caption{Performance comparison between original and augmented datasets.}
    \label{fig:learning_curves}
    \end{figure}

To quantify the benefits of the proposed data augmentation strategy, this paper first trained the model on the original, imbalanced dataset. The performance, averaged across five folds, was modest, achieving a Macro F1-score of 0.3540, an MAE of 0.7244, and a Quadratic Kappa of 0.2410. These results underscore the significant challenge posed by the inherent class imbalance and data scarcity in the high-risk categories.

\begin{figure}[t]
\centering
\includegraphics[width=0.7\linewidth]{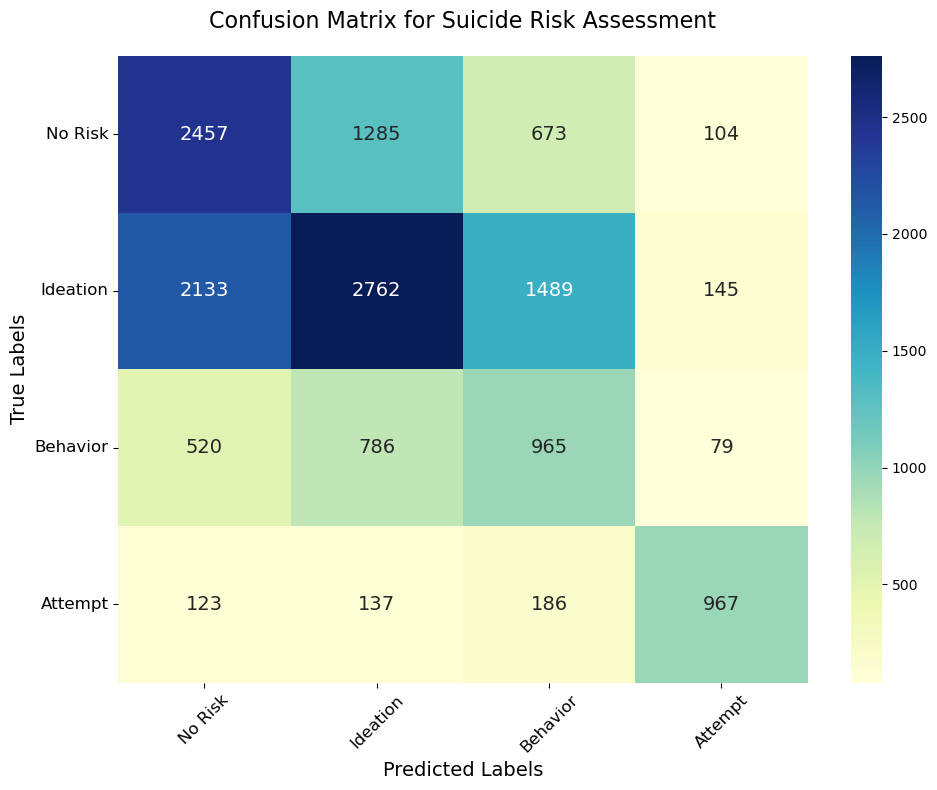} 
\caption{Confusion matrix on the out-of-fold validation set, illustrating class-wise prediction performance of the proposed model on the augmented data.}
\label{fig:confusion_matrix}
\end{figure}

Subsequently, this paper trained the model on the augmented dataset, which incorporates both in-sample LLM-generated variants and externally sourced data. The performance improved dramatically. The overall out-of-fold (OOF) Macro F1-score increased to 0.5098, representing a relative improvement of over 44\%. Concurrently, the MAE decreased to 0.6474, indicating a reduction in ordinal prediction error, and the QWK nearly doubled to 0.4692, signifying a much stronger agreement with the ground-truth risk levels. Figure~\ref{fig:learning_curves} illustrates the training dynamics, showing stable convergence of key metrics over epochs. The confusion matrix in Figure~\ref{fig:confusion_matrix} further details the model's predictive performance across the four risk levels on the validation set, demonstrating improved accuracy, particularly for the previously underrepresented classes. These results validate that the dual-pronged augmentation approach effectively mitigates class imbalance and enhances model generalization.

\begin{figure}[t]
\centering
\includegraphics[width=0.9\linewidth]{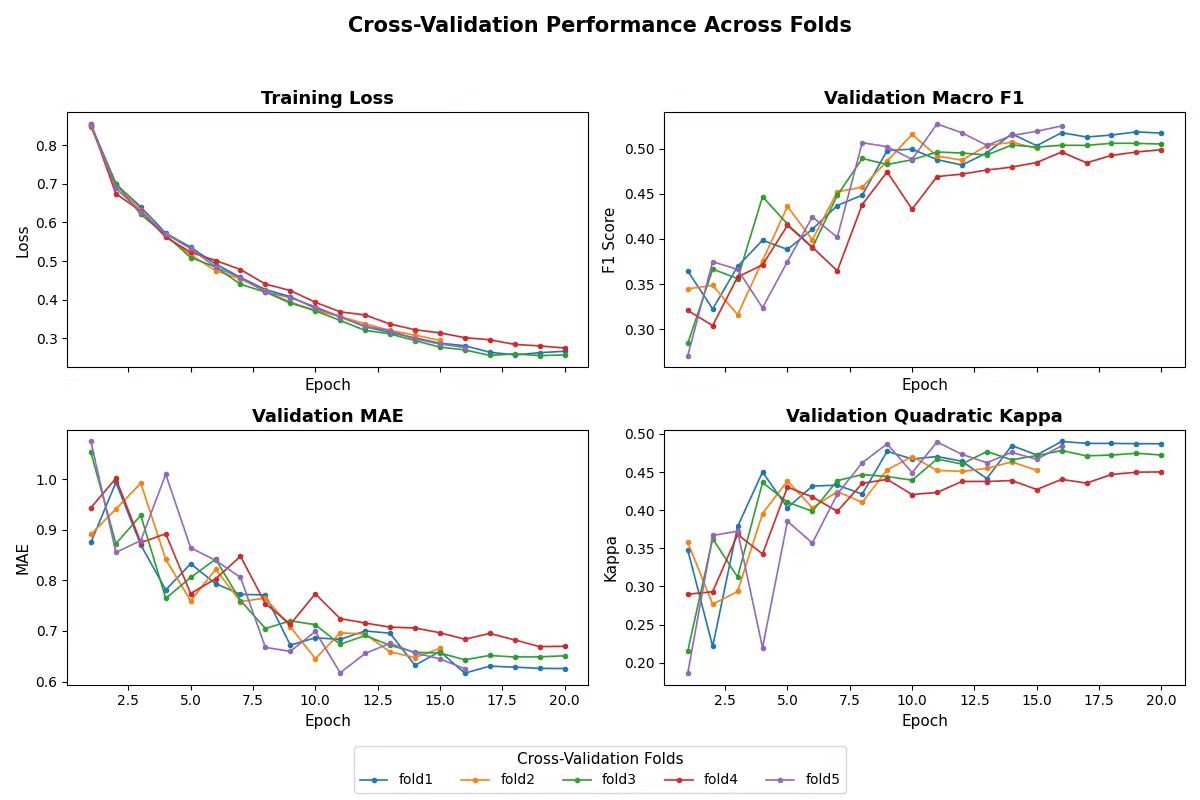} 
\caption{Training dynamics on the augmented dataset, showing the convergence of Macro F1, MAE, and Kappa over training epochs.}
\label{fig:learning_curves}
\end{figure}

\subsection{Comparison with Baseline Models}
\label{subsec:exp_baselines}

We benchmark the proposed model against two strong hierarchical baselines designed for similar text sequence classification tasks. The first baseline, \textbf{BiLSTM-MTL}, employs a hierarchical Bi-directional LSTM network for sequence modeling with a multi-task learning objective. The second, \textbf{Transformer-HAN}, is a Hierarchical Attention Network built upon Transformer blocks to capture dependencies at both post and sequence levels. For a fair comparison, all models were trained and evaluated on the augmented dataset.

The results, summarized in Table~\ref{tab:baseline_comparison}, demonstrate the superiority of the proposed architecture. The proposed model achieves a Macro F1-score of 0.5091, Quadratic Kappa of 0.4692, and Weighted F1-score of 0.4870, outperforming both the BiLSTM-MTL and the more advanced Transformer-HAN across most metrics. This superior performance highlights the effectiveness of the dual-head design, which synergistically combines ordinal and categorical predictions, along with the sophisticated temporal modeling and the tri-objective loss function tailored to the specific challenges of suicide risk assessment.

\begin{table}[h]
\centering
\caption{Comparison with Baseline Models on Performance Metrics}
\label{tab:baseline_comparison}
\begin{tabular}{lccc}
\toprule
\textbf{Model} & \textbf{Macro F1} & \textbf{Quadratic Kappa} & \textbf{Weighted F1} \\
\midrule
\rowcolor{green!10} BiLSTM-MTL & $\downarrow$ 0.4194 & $\downarrow$ 0.3419 & $\downarrow$ 0.4688 \\
\rowcolor{green!10} Transformer-HAN & $\downarrow$ 0.4906 & $\downarrow$ 0.4329 & $\uparrow$ 0.5028 \\
\rowcolor{blue!10} \textbf{Our Model} & \textbf{0.5091} & \textbf{0.4692} & \textbf{0.4870} \\
\bottomrule
\end{tabular}
\end{table}

\subsection{Ablation Study}
\label{subsec:exp_ablation}

To dissect the contribution of each key component in the proposed architecture, this paper conducted an extensive ablation study. This paper trained several variants of the model on the full augmented dataset for 8 epochs, systematically removing one component at a time: (1) \textbf{No Transformer}, where the 3-layer Transformer encoder is removed, and post embeddings are fed directly to the attention pooling layer; and (2) \textbf{No Features}, where the optional statistical feature fusion module is disabled.

The results, presented in Table~\ref{tab:ablation_study}, confirm that each component positively contributes to the model's overall performance. The full model serves as the baseline, achieving a Macro F1-score of 0.5193, MAE of 0.6446, and Kappa of 0.4575. The most substantial performance degradation occurred upon removing the Transformer encoder (\textbf{No Transformer}), with the F1-score dropping to 0.5020, MAE increasing to 0.6747, and Kappa decreasing to 0.4421. This underscores the critical role of the Transformer in capturing complex temporal dependencies between posts. Ablating the statistical features (\textbf{No Features}) resulted in slightly lower performance with F1-score of 0.5094, MAE of 0.6706, and Kappa of 0.4439, indicating that while textual information is dominant, behavioral features provide valuable complementary signals for risk assessment.

\begin{table}[t]
\centering
\caption{Ablation Study: Impact of removing key model components on performance metrics.}
\label{tab:ablation_study}
\begin{tabular}{lccc}
\toprule
\textbf{Model Variant} & \textbf{Macro F1} & \textbf{MAE} & \textbf{Kappa} \\
\midrule
\rowcolor{green!10} No Transformer & $\downarrow$ 0.5020 & $\uparrow$ 0.6747 & $\downarrow$ 0.4421 \\
\rowcolor{green!10} No Features & $\downarrow$ 0.5094 & $\uparrow$ 0.6706 & $\downarrow$ 0.4439 \\
\rowcolor{blue!10} Full Model & 0.5193 & 0.6446 & 0.4575 \\
\bottomrule
\end{tabular}
\end{table}

\section{Conclusion}

This paper proposed a hierarchical dual-head model for suicide risk assessment. The model forms a sequence-level representation with post-level MentalRoBERTa encoding and explicit temporal embeddings, and predicts via a CORAL ordinal head and a categorical head optimized by a tri-objective loss (CORAL, label-smoothed cross-entropy, and focal). On the augmented dataset, the model achieves a Macro F1 of 0.5098 with reduced MAE and higher Quadratic Weighted Kappa. Ablation results confirm the necessity of the Transformer encoder and the usefulness of optional statistical features, and demonstrate the effectiveness of the data augmentation pipeline.

Extensive experiments demonstrate consistent gains from the proposed architecture and training strategy. On the augmented dataset, the proposed model achieves an out-of-fold Macro F1 of 0.5098 (over 44\% relative improvement versus the original dataset), reduced MAE, and a substantially higher quadratic weighted kappa, indicating better ordinal consistency. Ablations confirm the importance of temporal modeling, dual-head prediction, and the tri-objective loss.

\section*{Acknowledgment}

We would like to express our sincere gratitude to the chair of the conference and organizing committee for the IEEE BigData 2025 Cup: Detection of Suicide Risk on Social Media for their dedicated organizational efforts. We also extend our thanks to our supervisors and classmates for their support. Additionally, we are grateful to all team members for their valuable contributions and collaboration throughout this project.

\balance
\bibliographystyle{IEEEtran}
\bibliography{main}

\section*{Appendix: Data Augmentation Prompts}

This appendix provides the detailed prompts and methodologies used for data augmentation in this study. The augmentation process employed two complementary approaches to address class imbalance and enhance model generalization.

\subsection*{Method 1: In-Sample Data Augmentation}

The first augmentation method leverages large language models (LLMs) to generate semantically equivalent variants of existing samples. The following prompt was used to guide the LLM in generating augmented data:

\textbf{Task Description:} You are an AI expert proficient in Natural Language Processing (NLP) and data augmentation. Your task is to generate high-quality augmented data for a study on suicide risk prediction. You must exercise extreme caution to ensure that the semantics and risk level of the generated text remain consistent with the original.

\textbf{Task:} I will provide a Python list containing 5 English posts. For each post in the list, you need to generate 3 new versions of the text that differ in expression but retain the exact same core meaning, emotional tone, and risk level.

\textbf{Data Augmentation Methods:} When generating new text, comprehensively apply the following two methods:
\begin{enumerate}
\item \textbf{Paraphrasing \& Synonym Replacement:} Use different vocabulary, phrases, or sentence structures to rephrase the original post.
\item \textbf{Simulated Back-Translation:} Imagine translating the text into another language and then back into English—preserve the core meaning but alter the wording and syntax. You do not need to perform actual translation; simply simulate this style of generation.
\end{enumerate}

\textbf{Core Rules (Must Strictly Adhere):}
\begin{itemize}
\item \textbf{Preserve Core Intent:} This is the most important rule. The augmented text must retain the emotion, level of despair, and severity of suicide risk (e.g., from ideation to planning to behavior) expressed in the original post. Do not dilute, distort, or alter the original intent.
\item \textbf{Maintain First-Person Perspective:} All newly generated posts must use the same first-person perspective ("I") as the original.
\item \textbf{Do Not Add New Information:} Do not fabricate details absent from the original text (e.g., specific locations, people, methods).
\item \textbf{Generate 3 Variants:} For each original post, you must provide exactly 3 distinct augmented versions.
\item \textbf{Strict Output Format:} Your final output must be a single, directly Python-parsable list. This list should contain 3 inner lists, each corresponding to the original data and containing 5 strings.
\end{itemize}

\textbf{Format Example:} If the input is ["Post A", "Post B", ...], your output format must be:
\begin{itemize}
\item \textbf{List 1:} ["Augmented version 1 for Post A", "Augmented version 1 for Post B", ..., "Augmented version 1 for Post E"]
\item \textbf{List 2:} ["Augmented version 2 for Post A", "Augmented version 2 for Post B", ..., "Augmented version 2 for Post E"]  
\item \textbf{List 3:} ["Augmented version 3 for Post A", "Augmented version 3 for Post B", ..., "Augmented version 3 for Post E"]
\end{itemize}

\subsection*{Method 2: External Data Acquisition and LLM-Assisted Labeling}

The second augmentation method involves collecting real-world posts from the Reddit subreddit r/suicidewatch using ethical web crawling techniques. The process includes:

\textbf{Data Collection Process:}
\begin{enumerate}
\item Crawl real posts from the Reddit subreddit suicidewatch
\item Collect user IDs, post text content, and post publication timestamps
\item Aggregate posts by user ID and sort by post timestamp
\item Group every 5 consecutive posts as a post\_sequence with timestamps
\item Use the 6th post as the reference for labeling the preceding 5 posts
\end{enumerate}

\textbf{Annotation Guidelines:} The annotation process follows the framework established by Li et al. (2022) with the following classification standards:

\textbf{0. Suicide Indicator (Mark as 0):} No suicidal risk in the post. Includes:
\begin{itemize}
\item Discussions of third parties' risks (e.g., friends' attempts/ideations, or others claiming the author has risks)
\item No desire to die at all
\end{itemize}

\textbf{1. Suicide Ideation (Mark as 1):} Focuses on thoughts/feelings without actions. Includes passive/active desires to die (e.g., "want to die," "end my life"). Also covers:
\begin{itemize}
\item Unrealistic methods (e.g., "I run towards a cop with a knife")
\item Passive thoughts despite denial (e.g., "I really don't want to die, but I don't know the value of my life")
\item Conditional thoughts (e.g., "If I had a gun, I would kill myself")
\end{itemize}

\textbf{2. Suicidal Behavior (Mark as 2):} Includes preparatory acts/plans toward a future attempt or self-inflicted violence without clear intent to die (e.g., self-harm). Subtypes:
\begin{itemize}
\item \textbf{Behavior-how:} Thinking about methods/tools (e.g., "I plan to buy a rope," searching for ways, self-harm like "I've started cutting myself")
\item \textbf{Behavior-plan:} Writing suicide notes, general plans (e.g., "I already got my suicidal notes ready")
\item \textbf{Behavior-when:} Planning at specific/fuzzy times (e.g., "I am going to kill myself soon")
\end{itemize}

\textbf{3. Suicidal Attempt (Mark as 3):} Refers to previous suicide attempts, defined as self-inflicted, potentially injurious behavior with intent to die that failed. Annotate if the post mentions past attempts (e.g., "I tried to kill myself," "I attempted suicide"). Annotate regardless of current suicidal thoughts.

\textbf{4. Irrelevant Text (Mark as 4):} Content completely unrelated to suicide. No mention of suicide attempts, behaviors, ideations, or third-party risks (e.g., posts about everyday activities, hobbies, or unrelated topics like "I went hiking today").

\textbf{LLM-Assisted Labeling Prompt:} Help me complete the text classification task regarding suicide risk. The following are five classification standards. [Classification standards as described above] Now please help me determine the risk level of the following 10 sentences. Send me the risk levels in a sequential format. Example: 1,2,3,0,4,1,2,3,4,3. Here are the sentences:

\textbf{Multi-LLM Ensemble Voting:} To ensure annotation quality and minimize individual model bias, we employ multiple large language models to independently annotate each sequence using the same prompt and classification guidelines. Each LLM provides its predicted risk level for the sixth post based on the unlabeled post group(6 posts). The final label is determined through a majority priority rule: the risk level that receives the highest number of votes across all LLMs is selected as the ground-truth label. In cases of ties, priority is given to the higher risk level to err on the side of caution in clinical safety considerations. This ensemble approach significantly improves annotation consistency and reliability compared to single-model labeling.

The combination of both augmentation methods resulted in a more balanced dataset that preserves the ordinal and categorical nuances of risk levels while significantly improving model generalization and robustness.

\end{document}